\documentclass[letterpaper]{article}
\usepackage{uai2020}
\usepackage[margin=1in]{geometry}

\usepackage{times}

\usepackage[utf8]{inputenc} 
\usepackage[T1]{fontenc}    
\usepackage{hyperref}       
\usepackage{url}            
\usepackage{booktabs}       
\usepackage{amsmath}
\usepackage{amsfonts}       
\usepackage{amssymb}
\usepackage{bm}
\usepackage{nicefrac}       
\usepackage{microtype}      
\usepackage{todonotes}
\usepackage[symbol]{footmisc}

\usepackage{printlen}

\usepackage{color}
\usepackage{graphicx}
\usepackage[noabbrev, capitalise]{cleveref}
\usepackage{amsthm}
\usepackage{placeins}
\usepackage{subcaption}
\usepackage{adjustbox}

\usepackage[round]{natbib}

\newcommand*{\cov}{\operatorname{Cov}}
\newcommand*{\Var}{\operatorname{Var}}

\newcommand*{\expectation}{\mathbb{E}}
\newcommand*{\tr}{\operatorname{Tr}}

\newcommand*{\dataset}[1]{\textit{#1}}

\newcommand*{\eg}{e.g., }
\newcommand*{\ie}{i.e., }

\newcommand*{\minibatchsum}{\sum_{b\in\mathcal{B}}}
\newcommand*{\gradientestimator}{\hat{g}_{bi}(\vepsilon_b)}
\newcommand*{\controlvariateterm}{\vc_{bi}\transpose\mycvtype(\vepsilon_b)} 

\newcommand*{\vzero}{\mathbf{0}}
\newcommand*{\identity}{\mathrm{I}}

\newcommand*{\transpose}{^\top}

\newcommand*{\newtext}[1]{#1}

\newcommand*{\cL}{\mathcal{L}}
\newcommand*{\params}{\boldsymbol{\theta}}
\newcommand*{\vc}{\mathbf{c}}

\newcommand*{\vw}{\mathbf{w}}
\newcommand*{\vW}{\mathbf{W}}
\newcommand*{\vG}{\mathbf{G}}
\newcommand*{\vepsilon}{\boldsymbol{\epsilon}}

\newcommand*{\myeqnstyle}{} 
\newcommand*{\mysumlimits}{} 
\newcommand*{\mycvtype}{\hat{\vw}}  

\title{Amortized variance reduction for doubly stochastic objectives}
\author{ {\bf Ayman Boustati \thanks{~~Work done while at Prowler.io.}} \\
University of Warwick \\
Coventry, UK
\And
{\bf Sattar Vakili}  \\
Prowler.io      \\
Cambridge, UK
\And
{\bf James Hensman}   \\
Prowler.io \\
Cambridge, UK
\And
{\bf ST John} \\
Prowler.io \\
Cambridge, UK
}

\begin{document}
\maketitle

\begin{abstract}
  Approximate inference in complex probabilistic models such as deep Gaussian processes requires the optimisation of doubly stochastic objective functions. These objectives incorporate randomness both from mini-batch subsampling of the data and from Monte Carlo estimation of expectations. If the gradient variance is high, the stochastic optimisation problem becomes difficult with a slow rate of convergence. Control variates can be used to reduce the variance, but past approaches do not take into account how mini-batch stochasticity affects sampling stochasticity, resulting in sub-optimal variance reduction. We propose a new approach in which we use a recognition network to cheaply approximate the optimal control variate for each mini-batch, with no additional model gradient computations. We illustrate the properties of this proposal and test its performance on logistic regression and deep Gaussian processes.
\end{abstract}

\section{INTRODUCTION}
\begin{figure}[t]
  \centering
  \includegraphics[width=\columnwidth]{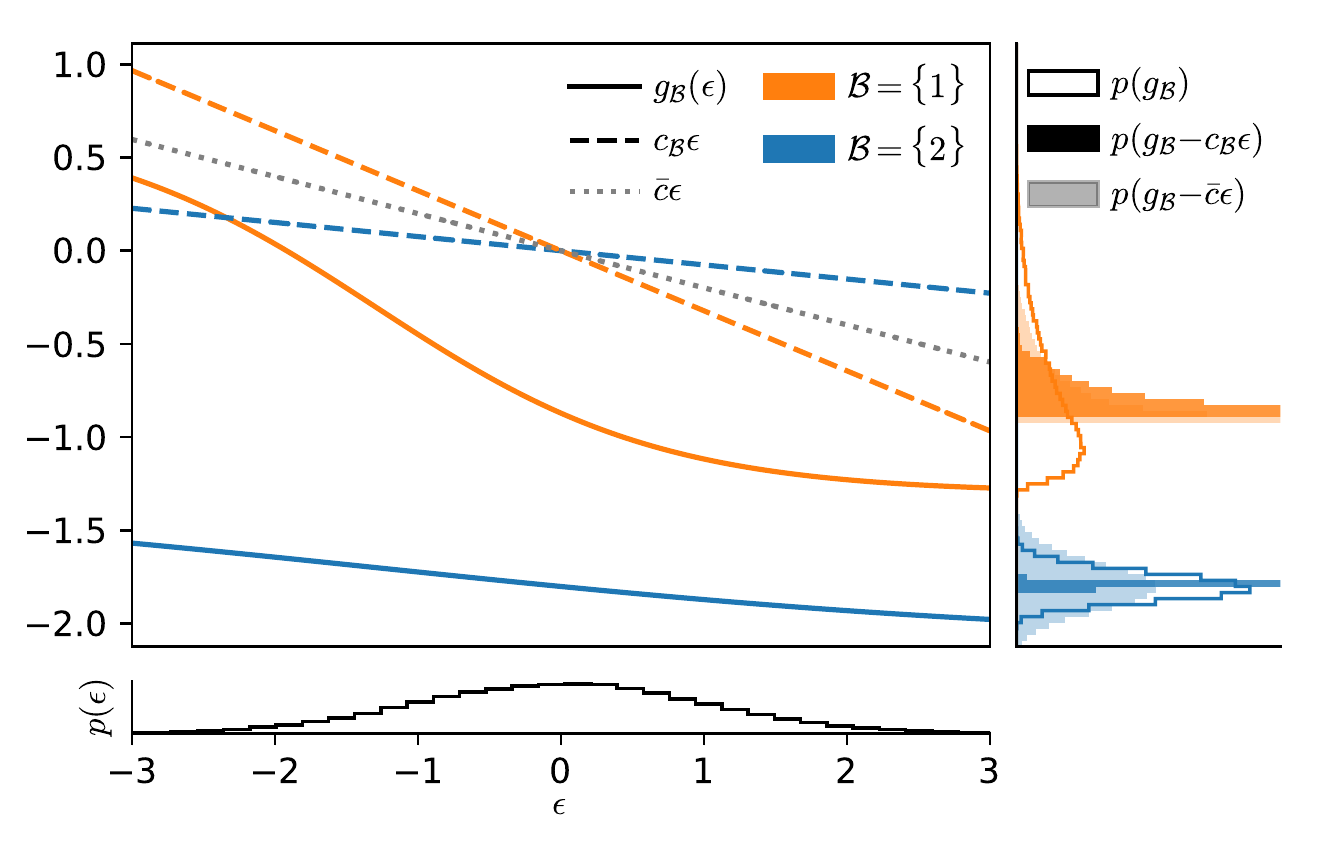}
  \vspace{-2em}
  \caption{In reparameterized variational inference, the gradient value $g$ is a function of the randomness sample $\epsilon \sim p(\epsilon)$. This functional relationship $g_\mathcal{B}(\epsilon)$ (solid lines) depends on the mini-batch $\mathcal{B}$ (orange vs blue).
  Here we show linear control variates (CVs) with batch-dependent coefficients $c_\mathcal{B}\epsilon$ (dashed lines) and the best batch-independent CV $\bar{c}\epsilon$ (dotted grey line). The right-hand plot shows the distribution of these expectation estimators for each mini-batch: without CV (outline), with batch-independent CV (shaded), and with batch-dependent CV (filled). The batch-dependent CV significantly reduces the variance, whereas here the batch-independent CV actually increases the variance for the blue mini-batch.}
  \label{fig:illustration}
\end{figure}

Many machine learning tasks such as regression and classification can be cast into a form in which we infer model parameters $\params$ by optimising an objective function $\cL = \sum_{n=1}^N \ell_n(\params)$ which is a sum over contributions from each data point $n$. We focus on objectives that contain an analytically intractable expectation, $\ell_n(\params) = \expectation_{p(\vepsilon)}[f_n(\vepsilon,\params)]$, such as in Black Box Variational Inference \citep{Ranganath2014bbvi}, Variational Auto-Encoders \citep{Kingma2014VAE}, or Deep Gaussian Processes \citep{Salimbeni2017dsdgp}.

In practice, such objectives are treated using Monte Carlo (MC) sampling to obtain an unbiased stochastic estimate of the expectation, $\hat{\ell}_n = \frac{1}{S} \sum_{s=1}^S f_n(\vepsilon_n^{(s)}, \params)$, where $\vepsilon_n^{(s)} \sim p(\vepsilon)$. We can then optimise using Stochastic Gradient Descent (SGD) on the noisy gradients \citep{robbins1951stochastic}. For large $N$, the evaluation of the full sum in $\cL$ is often computationally intractable. This can be addressed by subsampling mini-batches $\mathcal{B}\subset\{1,\dots,N\}$ of size $|\mathcal{B}|$ from the full data set, introducing additional noise and leading to a doubly stochastic objective function:
\begin{align}
    \label{eqn:doubly_stochastic1}
    \hat{\cL} := \frac{N}{|\mathcal{B}| S}\sum_{b\in\mathcal{B}}\sum\nolimits_{s=1}^S f_b(\vepsilon_b^{(s)}, \params) ,
\end{align}
with $\expectation [\hat{\cL}] = \cL$.

The variance of the gradients of $\cL$ affects both the rate of convergence of the optimisation and how close can the optimiser get to the optimum. This motivates various approaches for reducing either mini-batch variance (\eg \citet{Johnson2013svrg}) or the variance due to MC estimation of the expectation \citep{Ranganath2014bbvi, Roeder2017stickingthelanding}. A common approach for variance reduction are control variates (see \cref{sec:control_variates}), which have recently been adopted in the literature \citep{Paisley2012vbisearch, Miller2017reducingreparameterization, Grathwohl2018backpropagation, Geffner2018usinglargeensembles}. The focus for the latter work is on deriving and applying control variate schemes to MC objectives, specifically in the context of Variational Inference (VI).

However, to the best of our knowledge, the schemes in the literature do not consider the mini-batching case and do not explicitly take into account how the context of the data point $b$ affects the dependence of $f_b$ on $\vepsilon$.
This dependence is illustrated in \cref{fig:illustration} at the example of Bayesian logistic regression. The gradient $g_\mathcal{B}(\epsilon)$ as a function of the randomness $\epsilon$ of a doubly stochastic objective is shown for two different mini-batches $\mathcal{B}$. In this simplified case, each batch consists of a single context point. The two different context points induce different relationships between randomness and gradient value as shown by the solid lines. This means that the two gradients correlate differently with the randomness, resulting in different control variates represented by the dashed lines. For comparison, we include a batch-independent control variate for the expectation estimator (dotted line) which has to average over all contexts. Adapting the control variate to the batch significantly reduces the variance, shown in the right-hand panel in \cref{fig:illustration}.

In this work, we propose a novel idea for computing control variates that adapt to the context (mini-batch) of the controlled estimators (the gradient). The new formulation takes into account the dependence of the MC estimate on the data by using a recognition network to learn an adaptive control variate coefficient. We derive a low-variance objective function to train the network to approximate the optimal control variate coefficient per batch. Additionally, we propose two computationally cheaper alternatives to the network objective with higher variance. All control variate objectives re-use the already computed model objective gradient, and hence do not require extra back-propagation steps. We empirically test the properties of our proposed method in \cref{sec:experiments}.

\section{METHODOLOGY}

In \cref{sec:control_variates}, we start with a review of control variates and highlight the importance of computing the optimal control variate coefficient in the general case to allow for maximal variance reduction. We introduce the dependence of the gradients and the control variates on the selected mini-batch and propose learning context-aware control variate coefficients in \cref{sec:controlling} . Finally, in \cref{sec:training_recognition_network} we derive the objectives for the control variate coefficients that allow amortisation through a recognition network.

\subsection{CONTROL VARIATES}
\label{sec:control_variates}
We want to reduce the variance of an unbiased stochastic estimator $\hat{g}_\theta(\vepsilon)$\footnote{We use the $\hat{\phantom{m}}$ symbol (as well as $\tilde{\phantom{m}}$) on top of functions of random variables to denote the estimate of this function obtained by evaluating the relevant estimator. In the following we drop the dependence on $\theta$ to lighten the notation.} for an intractable expectation $\expectation[g(\vepsilon)]$, where $\vepsilon \sim p(\vepsilon)$ is a random variable. We consider a different function $w(\vepsilon)$ whose expectation is known analytically, $\expectation[w(\vepsilon)] = W$. Then $C (w(\vepsilon) - W)$ has zero expectation for any $C$, and its unbiased estimator, $C(\hat{w}(\vepsilon) - W)$, can be subtracted from the original estimator,
\begin{equation}
\label{eqn:basic_cv}
\tilde{g}(\vepsilon) = \hat{g}(\vepsilon) - C(\hat{w}(\vepsilon) - W) .
\end{equation}
This new estimator has the same expectation as the original estimator and is also unbiased. Minimising its variance $\Var[\tilde{g}]$ gives the optimal $C^* = \cov[\hat{g}, \hat{w}] / \Var[\hat{w}]$, and $\tilde{g}$ will have lower variance than $\hat{g}$ if $g(\vepsilon)$ and $w(\vepsilon)$ are correlated. In particular, choosing $C$ optimally results in variance reduction of
\begin{equation}
\label{eqn:variance-reduction}
\Var[\tilde{g}] = (1 - \rho_{g, w}) \Var[\hat{g}],
\end{equation}
where $\rho_{g, w}$ is the Pearson correlation coefficient between $g$ and $w$.
In practice, however, computing $C^*$ is not possible, as $\cov[\hat{g}, \hat{w}]$ and $\Var[\hat{w}]$ cannot be evaluated exactly, and are usually estimated from the optimisation statistics, e.g.\ running averages \citep{Paisley2012vbisearch}. Another option is to pre-specify $C$ and keeping it fixed \citep{Miller2017reducingreparameterization, Grathwohl2018backpropagation}.

Neither option is convincing for the doubly stochastic case. The first option has very high variance due to the presence of mini-batch stochasticity in addition to sampling stochasticity. The second option is unreliable as pre-specifying an arbitrary value for $C$ does not guarantee optimal variance reduction as can be seen in \eqref{eqn:variance-reduction}.

In \cref{sec:controlling} we will specify $C$ as a context-dependent adaptive parameter that is learned through the optimisation. In \cref{sec:training_recognition_network}, we discuss the corresponding training objectives for $C$.

\subsection{CONTROLLING MINI-BATCH GRADIENTS}
\label{sec:controlling}

For gradient-based optimisation we need the derivatives of the objective \eqref{eqn:doubly_stochastic1} with respect to the model parameters $\{\theta_i\}_{i=1}^P$. The estimated gradient contains a sum over mini-batch elements $b$,
\begin{equation}
    \label{eqn:gradient}
    \frac{\partial\hat{\mathcal{L}}}{\partial \theta_i} \propto \sum_{b\in\mathcal{B}}\frac{\partial f_b}{\partial\theta_i}(\vepsilon_b, \params) = \sum_{b\in\mathcal{B}}\hat{g}_{bi}(\vepsilon_b) =: \hat{G}_i ,
\end{equation}
\newtext{where we chose $S=1$ to simplify the equations (the extension to multiple MC samples is straightforward).}
Note that $\mathcal{B}$ is a random subset of $\{1,\dots,N\}$, \ie the $b$'s are indices into the full dataset, and each term gets its own realisation $\vepsilon_b$ of the randomness.
We want to improve the optimisation performance by reducing the variance of this gradient.
As demonstrated in \cref{fig:illustration}, each partial gradient estimator $\hat{g}_{bi}(\vepsilon_b)$ may have a different dependence on the randomness. To account for this, we introduce separate control variates for each term (data point) in the sum in \eqref{eqn:gradient}. For a single partial gradient, we define the controlled gradient estimator
\begin{equation}
\label{eqn:gradient_cv}
\tilde{g}_{bi}(\vepsilon_b) := \gradientestimator - \controlvariateterm.
\end{equation}
\newtext{(Here and in the following section we subsume the analytic expectation into the definition of the control variate such that $\hat{\vw}(\vepsilon)$ already has zero mean. We use the same type of control variates for all parameters; in principle, we could have a different $\hat{\vw}_i$ per parameter $\theta_i$. Note that in any case there are per-parameter coefficients $\vc_{bi}$.)}
In general, the mapping $\mycvtype(\vepsilon)$ may have a different number of components than the randomness $\vepsilon$ itself. For simplicity, in the following we assume both $\vepsilon$ and $\mycvtype(\vepsilon)$ are $D$-dimensional. Note that $\mycvtype(\cdot)$ does not depend on the batch element $b$; the dependence is captured in the coefficients $\vc_{bi}$, which is a vector of length $D$ for each index pair $b,i$.

Specifying the problem this way allows us to explicitly model each control variate coefficient per data point. Under this setting, the new estimator for the gradient is
\begin{equation}
    \label{eqn:controlled_estimator}
    \tilde{G}_i = \minibatchsum \tilde{g}_{bi}(\vepsilon_b) = \minibatchsum (\gradientestimator - \controlvariateterm).
\end{equation}
The control variate coefficients $\vc_{bi}$ can be set to optimally reduce the variance of $\tilde{G}_i$ by solving
\begin{equation}
    \min_C \tr(\cov[\tilde{\vG}]),
\end{equation}
where $C$ is the collection of $\vc_{bi}$ and has shape $N \times P \times D$, as we need separate coefficients for all $N$ data points.

Computing and storing these can be computationally prohibitive for large data sets, hence we propose to amortise the cost of this computation by using a recognition network $r_\phi: \mathcal{Y} \rightarrow \mathbb{R}^{P \times D}$ that outputs the coefficients for each mini-batch throughout the optimisation, where
\begin{equation}
    \label{eqn:recognition_network}
    \vc_{bi} = [r_{\phi}(y_b)]_i
\end{equation}
is a vector of dimension $D$ and $y_b \in \mathcal{Y}$ are context points (\eg feature vector and target for the $b$th data point in a supervised learning problem) and $\phi$ are the recognition network parameters.

As the control variate only adds terms to the gradients of the model's optimisation objective that are zero in expectation, we do not change the minima of the objective. This means that the extra parameters of the recognition network will \emph{not} lead to overfitting.

\subsection{TRAINING THE RECOGNITION NETWORK}
\label{sec:training_recognition_network}
Intuitively, we require the recognition network $r_\phi(\cdot)$ to output coefficients that minimise the variance of the controlled gradient estimator \eqref{eqn:controlled_estimator}. This gives the training objective for the parameters $\phi$:
\begin{equation}
    \label{eqn:cv_objective}
    \myeqnstyle
    \min_\phi \tr{(\cov[\tilde{\vG}}]) = \min_\phi \sum\mysumlimits_{i=1}^P\Var[\tilde{G}_i].
\end{equation}
The $i$th term in the sum in \eqref{eqn:cv_objective} is
\begin{align}
    &\myeqnstyle
    \Var[\tilde{G}_i] = \Var\Big[\minibatchsum\big(\gradientestimator - \controlvariateterm\big)\Big]
    \nonumber\\
    &\quad\myeqnstyle
    = \minibatchsum \Var\big[\gradientestimator - \controlvariateterm\big]
    \nonumber\\
    &\quad\myeqnstyle
    = \minibatchsum \Big(
            \Var\big[\gradientestimator] + \Var\big[\controlvariateterm\big]
        \nonumber\\
    &\qquad
            - 2 \cov\big[\gradientestimator, \controlvariateterm\big]
    \Big)
    \nonumber\\
    &\quad\myeqnstyle
    = \text{const} + \minibatchsum\Big(
        \expectation\big[(\controlvariateterm)^2\big]
    \nonumber\\
    &\qquad
        - 2 \expectation\big[(\gradientestimator)(\controlvariateterm)\big]
    \Big) ,
    \label{eqn:partial_grads_objective}
\end{align}
and we discard the terms that do not contain $\vc_{bi}$ and hence do not give gradients for $\phi$.
For most problems, the expectations are intractable; we estimate these with MC sampling and define
\begin{gather}
    \myeqnstyle
    \tilde{V}_i = \minibatchsum \big((\controlvariateterm)^2 
    - 2 (\gradientestimator)(\controlvariateterm)\big) .
\label{eqn:partial_grads_objective1}
\end{gather}
We can now learn the optimal recognition network parameters $\phi$ using SGD (or variants) on
\begin{equation}
    \label{eqn:partial_grads_objective2}
    \myeqnstyle
    \min_\phi \sum\mysumlimits_{i=1}^P \tilde{V}_i .
\end{equation}
To train the parameters $\phi$, we need to compute gradients of $\sum_i \tilde{V}_i$. Examining the chain rule around the outputs of the recognition network, $ \frac{\partial \tilde{V}_i}{\partial c_{bid}} \frac{\partial c_{bid}}{\partial \phi} $, the second term is computed by backpropagation through the network, and the cost of computing the first term depends on the form of the estimator $\tilde{V}_i$. 

The recognition network objective using  \eqref{eqn:partial_grads_objective1} requires the partial gradients \emph{per data point} $\hat{g}_{bi}(\vepsilon_b)$ of the original objective function, and we call this the \textbf{partial gradients estimator}. In common reverse-mode automatic differentiation libraries \newtext{such as TensorFlow and PyTorch}, it requires $|\mathcal{B}|$ additional backward passes on the model objective, each at least $O(|\mathcal{B}|)$, so this becomes prohibitively expensive when the mini-batch size is large.
To overcome this limitation \newtext{in current implementations}, we derive two further estimators for the recognition network objective that are computationally cheaper, albeit with higher variance.

\subsubsection{The Gradient Sum Estimator}
To avoid the partial gradients in \eqref{eqn:partial_grads_objective1}, we return to the $i$th term of the sum in \eqref{eqn:cv_objective}. Instead of taking the sum out of the variance, we separate the sum over partial gradients from the control variates:
\begin{align}
    \Var[\tilde{G}_i] &\myeqnstyle = \Var\big[ \minibatchsum \gradientestimator - \minibatchsum \controlvariateterm \big]
    \nonumber\\
    &\myeqnstyle = \Var\big[ \hat{G}_{i} - \minibatchsum \controlvariateterm \big] .
    \label{eqn:start_from}
\end{align}
We can expand the variance of a sum of two terms as
\begin{align}
\myeqnstyle
    \Var[\tilde{G}_i] &= \myeqnstyle \Var[\hat{G}_{i}] + \Var\big[ \minibatchsum \controlvariateterm \big] \nonumber\\
    \myeqnstyle
    &\quad\myeqnstyle - 2\cov\big[ \hat{G}_{i}, \minibatchsum \controlvariateterm \big] \nonumber\\
    &= \myeqnstyle \text{const} + \minibatchsum \Big( \expectation[(\controlvariateterm)^2]
     \nonumber\\
    &\quad - 2 \expectation[(\hat{G}_{i})(\controlvariateterm)] \Big) ,
    \label{eqn:gradient_sum_objective}
\end{align}
and by replacing the expectations with MC estimates, we arrive at a new estimator
\begin{gather}
    \tilde{V}^\text{GS}_i = \minibatchsum \big( (\controlvariateterm)^2  
    - 2 (\hat{G}_{i})(\controlvariateterm) \big) .
\end{gather}
This estimator is similar in form to the partial gradients estimator, replacing the gradient per data point with the sum over the whole mini-batch; we call this the \textbf{gradient sum} estimator. As it does not require any additional backward passes, it is much cheaper to compute. One can intuitively see that this estimator has a higher variance than the partial gradients estimator as it additionally includes cross terms that would be zero in expectation.

\subsubsection{The Squared Difference Estimator}
Alternatively, we can continue from \eqref{eqn:start_from} by expanding the variance into moment expectations:
\begin{gather*}
    \myeqnstyle
    \Var[\tilde{G}_{i}]
    = \expectation\big[\big(\hat{G}_{i} - \minibatchsum \controlvariateterm\big)^2\big] \\
    \myeqnstyle
    - \big(\expectation\big[\hat{G}_{i} - \minibatchsum \controlvariateterm \big]\big)^2 ,
\end{gather*}
where the control variate term has no contribution inside the second expectation by definition, and $\expectation[\hat{G}_i]$ is a constant with respect to the recognition network parameters $\phi$.
Evaluating the remaining expectation using MC gives us the \textbf{squared difference} estimator:
\begin{equation}
    \myeqnstyle
    \tilde{V}^\text{SD}_i = \big(\hat{G}_{i} - \minibatchsum \controlvariateterm \big)^2 ,
    \label{eqn:squared_difference}
\end{equation}
which is also cheap to compute. In contrast to $\tilde{V}^\text{GS}_i$, it includes the second moment of $\hat{G}_i$. This is similar to a regression problem that uses $\hat{w}_{d}(\vepsilon_b)$ as basis functions to learn the gradient $\hat{G}_i$.

\section{ILLUSTRATIVE EXAMPLE: CONTROL VARIATES FOR GAUSSIAN BASE RANDOMNESS}
\label{sec:gaussian_base_randomness}
So far our discussion has been general. To implement a control variate, we need to specify both the distribution of the base randomness $\vepsilon$ and the functional form of the control variate $\vw(\vepsilon_n)$. In principle, any functional form for control variates from the literature can be used with this method, e.g.\ \citet{Paisley2012vbisearch,Ranganath2014bbvi,Miller2017reducingreparameterization}.
For the sake of simplicity, we illustrate our proposal on a simpler control variate form for the special case of Gaussian base randomness, which is of direct interest to many applications in VI. 

We assume $\vepsilon \sim \mathcal{N}(\vzero, \identity_D)$ without loss of generality.\footnote{In the general case where $\vepsilon \sim \mathcal{N}(\boldsymbol{\mu}, \Sigma)$, we can simply apply the reparameterisation $\vepsilon = \boldsymbol{\mu} + \operatorname{Cholesky}(\Sigma) \vepsilon_0$ with $\vepsilon_0 \sim \mathcal{N}(\vzero, \identity_D)$.}
In this section we introduce explicit forms for $\vw(\vepsilon_n)$ for this case, starting with linear control variates, and then extend the discussion to higher-order polynomials.

\subsection{LINEAR GAUSSIAN CONTROL VARIATES}
The simplest control variate is an element-wise linear function of $\vepsilon_n$,
\newcommand*{\valpha}{\boldsymbol{\alpha}}
\newcommand*{\vbeta}{\boldsymbol{\beta}}
\begin{equation}
    \vw(\vepsilon_n) = \valpha + \vbeta \circ \vepsilon_n , \vepsilon_n \sim \mathcal{N}(\vzero, \identity_D),
\end{equation}
with $\circ$ representing the element-wise product. Its expectation is $\vW = \expectation[\vw(\vepsilon_n)] = \valpha$, and the control variate simplifies to $\vbeta \circ \vepsilon_n$. We can also absorb $\vbeta$ into the control variate coefficient $\vc_{ni}$, which results in the following controlled version of the gradient component $i$, for data point $n$:
\begin{equation}
\label{eqn:linear_cv}
\tilde{g}_{ni}(\vepsilon_n) = \hat{g}_{ni}(\vepsilon_n) - \vc_{ni}\transpose\vepsilon_n .
\end{equation}
Intuitively, one can think of control variates of this form as injecting the estimator with information on the linear dependence of the gradient on the noise. To understand this further, we take a look at the first-order Taylor expansion of the gradient component $g_{ni}(\vepsilon_n)$ around $\vepsilon_n = \mathbf{0}$,
\begin{equation}
\label{eqn:linear_jacobian}
g_{ni}(\vepsilon_n) = g_{ni}(\mathbf{0}) + \nabla g_{ni}(\mathbf{0})\transpose \vepsilon_n + O(\vepsilon_n^2) .
\end{equation}
If the gradient is sufficiently linear with respect to $\vepsilon_n$ (\ie the $O(\vepsilon_n^2)$ terms are negligible), and when $\vc_{ni}$ is a good approximation to the Jacobian at $\mathbf{0}$, the estimator in (\ref{eqn:linear_cv}) will have low variance.

\subsection{HIGHER-ORDER POLYNOMIALS}
In general, the gradient is unlikely to be linear with respect to the noise, especially for complicated models and objectives. To overcome this, we can use higher-order polynomials to capture some of the non-linear dependence of the gradient on the noise. Consider the following form for $\vw(\vepsilon_n)$:
\begin{equation}
    \myeqnstyle
    \vw(\vepsilon_n) = \sum_{k=1}^K \valpha_{k} \circ \vepsilon_n^k ,
\end{equation}
where the $k$th power is evaluated element-wise. $\vW$ can be easily computed and would correspond to the sum of diagonal parts of the first $K$ moment tensors of the multivariate Gaussian distribution, scaled by $\valpha_{k}$. For instance, for $K=2$ the control variate is given by
\begin{equation}
    \label{eqn:quadratic_cv}
    \valpha_{1} \circ \vepsilon_n + \valpha_{2} \circ (\vepsilon^2_n - \text{diag}(\identity_D)) .
\end{equation}
We can again simplify  by absorbing the $\valpha_{k}$ into the control variate coefficient, with slight adjustments to the controlled gradient estimator.

We make the following observation:
\paragraph{Remark 1:} A linear combination of control variates is also a valid control variate, \ie $\tilde{g}_{ni}(\vepsilon_n) = \hat{g}_{ni}(\vepsilon_n) - \sum_{k=1}^K (\vc_{ni}^{(k)})\transpose (\hat{\vw}_k(\vepsilon) - \vW_k)$ is unbiased.
By considering each term in (\ref{eqn:quadratic_cv}) as a separate control variate, we can write the $i$th component of the controlled gradient at $n$ as
\begin{equation}
\label{eqn:quadratic_cv2}
\tilde{g}_{ni}(\vepsilon_n) = \hat{g}_{ni}(\vepsilon_n) - (\vc_{ni}^{(1)})\transpose\vepsilon_n - (\vc_{ni}^{(2)})\transpose(\vepsilon_n^2 - \text{diag}(\identity_D)) .
\end{equation}
The same construction trivially extends to $K>2$.

\subsection{BRIEF DISCUSSION}
The simple examples of the linear and polynomial control variates presented above illustrate the importance of choosing a good control variate coefficient $C$. For instance, in the linear case in \eqref{eqn:linear_cv} the control variate function $\vw(\vepsilon_n) = \vepsilon_n$ does not provide any extra information on the estimator on its own, since we are essentially just adding noise to the MC estimate. However, with the selection of a good control variate coefficient $\vc_{n}$ for data point $n$, we introduce structure to the noise that contains information about the behaviour of the controlled quantity with respect to the Gaussian noise in the form of the Jacobian in \eqref{eqn:linear_jacobian}. Indeed the optimal coefficient $\vc^*_{n}$ for the linear control variate contains the Jacobian term.

\section{RELATED WORK}
Control variates are widely used to reduce the gradient variance of stochastic objectives, mainly motivated by VI. A comprehensive review can be found in \citet{Geffner2018usinglargeensembles}. Here, we highlight some relevant work and compare it to our contribution.

\citet{Paisley2012vbisearch} first introduce the idea of using control variates to reduce the gradient variance in VI. They propose using a bound on the objective or an approximation of the model as control variates. \citet{Ranganath2014bbvi} build on this work, using the score function of the approximate posterior to control the gradient of Black Box Variational Inference objectives.

Inspiration for our work comes from \citet{Grathwohl2018backpropagation}, where they use a recognition network to approximate the model and its gradient as a control variate. \citet{Miller2017reducingreparameterization} derive an approximation to the reparameterisation gradient for Gaussian variational distributions by performing a first-order Taylor expansion of the gradient, using this approximation as a control variate. Our work is related to this construction where the recognition network can be viewed as a cheap approximation to the linear term in the Taylor expansion of the gradient (\ie the Hessian of the model objective) in the case of the linear construction of \cref{sec:gaussian_base_randomness}. 

The unifying work of \citet{Geffner2018usinglargeensembles} categorises different control variate schemes for VI objectives. Additionally, they propose combining them to achieve greater variance reduction. They derive an optimal rule for this combination based on Bayesian risk minimisation.

These related works do not consider the effect of mini-batching on the proposed control variates. Our work should be viewed as \emph{complementary} to many of the methods mentioned above. Indeed, \citet{Geffner2018usinglargeensembles} show that a combination of control variates is usually more desirable that a single scheme. The method we proposed can be considered an extra addition to the control variate toolkit for doubly stochastic objectives, to take the effect of mini-batch stochasticity on the control variates into account. Our method can also be combined with other variance reduction methods such as extra sampling.

\section{EXPERIMENTS}
\label{sec:experiments}

\begin{figure*}[t!]
    \centering
    \begin{subfigure}{0.95\textwidth}
        \vspace{0.1in}
        \centering
        \includegraphics[width=\linewidth]{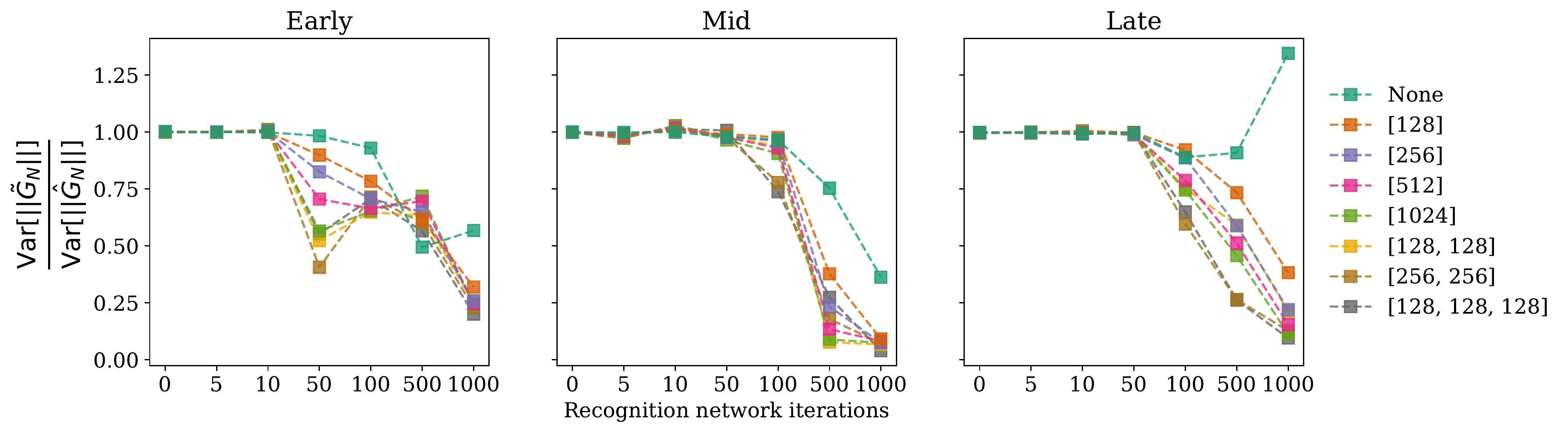}
        \caption{Logistic regression on \textit{titanic}.}
    \end{subfigure}
    \begin{subfigure}{0.95\textwidth}
        \vspace{0.1in}
        \centering
        \includegraphics[width=\linewidth]{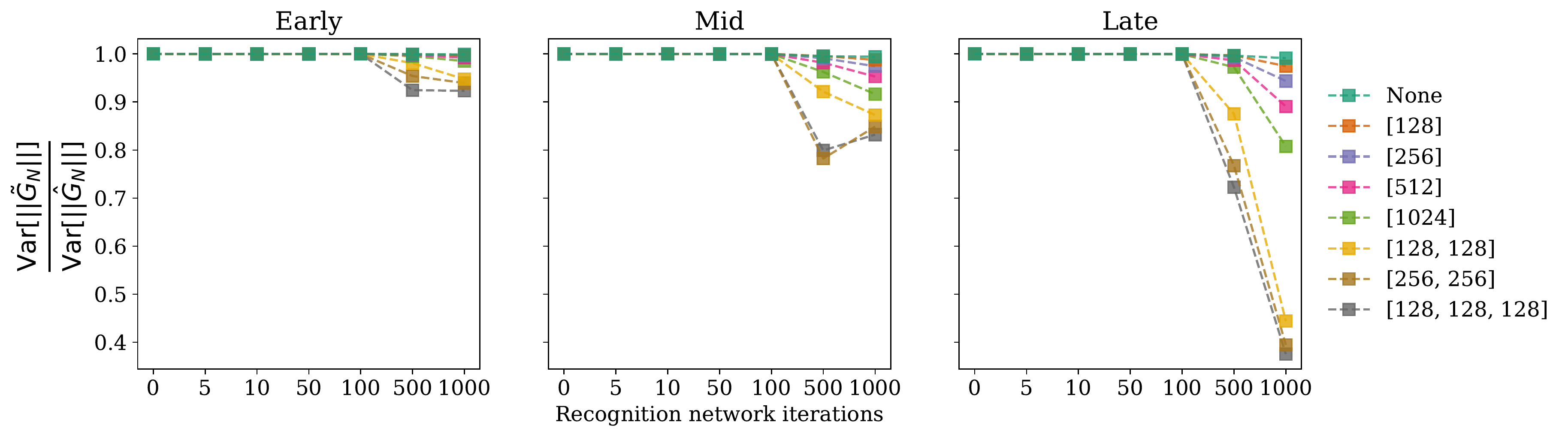}
        \caption{DGP on \textit{airfoil}.}
    \end{subfigure}
    \caption{Variance reduction at different points in the objective optimisation (lower is better); early: 10 steps, mid: 200 steps, late: 1000 steps. The results are shown for the \emph{linear} control variate from \cref{sec:gaussian_base_randomness}. Recognition network training uses the \emph{squared difference} objective optimised with Adam with learning rate of $10^{-2}$ for the logistic regression and $10^{-3}$ for the DGP. For a small number of iterations on the recognition network, it struggles to learn a good control variate coefficient. Continuing the network optimisation, it is able to learn good control variate coefficients that significantly reduce the variance in comparison to the context-free coefficient. Also notable is that the variance reduction is more pronounced at the later stages of the model optimisation.}
    \label{fig:static_variance_reducion}
\end{figure*}

Our discussion thus far applied to the general class of doubly stochastic objectives. For our experiments we focus on objectives arising from VI problems. Amortising the computation of the control variate coefficients in this setting is advantageous since context arises naturally from the data in the underlying models. 

In this section, we aim to answer three questions: a) To what extent can amortising with a recognition network reduce the variance compared to a fixed context-free control variate coefficient? b) How well can we train the recognition network in an online setting? c) What difference can an amortised control variate make in practice?

\subsection{SETUP}
We investigate (a), (b) and (c) on a classification task on the \dataset{titanic} dataset using a Bayesian logistic regression model and on a regression task on the \dataset{airfoil} dataset using a Deep Gaussian Process (DGP). 

For the Bayesian logistic regression model, we use the reparameterisation gradient formulation of the VI problem. We choose a Gaussian approximate posterior, where we learn the mean vector and the full covariance matrix. We select a unit Gaussian prior on the weights.

For the DGP model, we use a 2-layer model with inner layer dimension of 5, and a Squared Exponential kernel for the GP priors. We use the doubly stochastic formulation of the VI problem \citep{Salimbeni2017dsdgp}. We learn the parameters of the approximate Gaussian posterior, keeping the hyperparameters fixed. The inducing locations are fixed and selected as the centroids of $k$-means clusters from the data.

Throughout, we use Adam \citep{Kingma2015Adam} for both the model objective function optimisation and the recognition network objective optimisation. We use a single-sample MC estimate of the gradients and control these when stated, applying the linear and quadratic control variates introduced in \cref{sec:gaussian_base_randomness}. We initialise the recognition network with Xavier initialisation \citep{glorot2010understanding} and use ReLU activations in the hidden layers.

We compare our proposal to a context-free control variate. In this instance, this is implemented as an optimisable quantity that does not depend on data and uses the same optimisation objectives (\eqref{eqn:gradient_sum_objective} \& \eqref{eqn:squared_difference}) as the recognition network, i.e. $\mathbf{c}$ is independent of the mini-batch $\mathcal{B}$ in these objectives. This is equivalent to approximating the coefficient with an exponentially weighted moving of the empirical covariance of the gradient and the control variate estimates.

\subsection{VERIFICATION OF VARIANCE REDUCTION}
\label{section:static_variance_reduction}
The first question considered is whether the recognition network has the capacity to amortise the control variate coefficients and how well it can learn these versus a context-free coefficient? To test this, we freeze the model parameters at three points in the optimisation -- early (10 steps), mid (200) steps, and late (1000 steps) -- then optimise the control variate coeficient only. For each period, we iteratively sample a gradient value then perform an optimisation step on the recognition network. We repeat this procedure for 1000 steps and record the variance reduction at different steps.
The variance reduction is measured by the ratio $\Var[\|\tilde{G}_N\|] / \Var[\|\hat{G}_N\|]$, where $\tilde{G}_N$ and $\hat{G}_N$ are the controlled and uncontrolled gradients, respectively, over the mini-batch (size 10), and $\|\cdot\|$ is the gradient norm. We compare different network sizes to see the effect this has on variance reduction.

\cref{fig:static_variance_reducion} shows that amortising the control variate coefficient computation induces greater variance reduction than optimising a context-free coefficient (labelled as \emph{None} in the figure). 
The variance reduction does not occur immediately, as the control variate coefficients need to be optimised in all cases to reduce the variance. Also notable is that the amount of variance reduction depends on the optimisation stage of the model; at later stages of the model optimisation, the variance reduction is more pronounced. This is likely a property of both the model and the control variate where the gradients in the beginning of the optimisation have more pronounced non-linearities with respect to the noise. This can also be seen in the amount of variance reduction in logistic regression compared to the DGP. The gradients in the logistic regression models are approximately linear with respect to the noise, while in the DGP gradients have a more complex dependency on the noise. Finally, we can see that the variance reduction potential depends on the capacity of the network, where wider and deeper networks learn better control variate coefficients. Deeper networks reduce the variance more strongly than wider networks, which correspond to a highly non-linear mapping from the context points to the control variate coefficient.

\subsection{SIMULTANEOUS OPTIMISATION OF OBJECTIVE FUNCTION AND CONTROL VARIATE COEFFICIENT}
\label{sec:dynamic_variance_reduction}

In practice, the recognition network needs to be able to learn the control variate coefficients while the model is being optimised, giving a moving target. In this section, we investigate the viability of chasing this target by simultaneously optimising the model objective and recognition network. We use a recognition network with three layers of size 128 each, as this architecture showed the largest variance reduction in \cref{section:static_variance_reduction}. In each step in the optimisation procedure, we compute one gradient estimate of the model objective for a mini-batch of size 10. We take one Adam step on the recognition network, then we apply the control variate correction to the sampled gradient and take an Adam step on the model parameters. We measure the variance of the gradient at different periods in the optimisation by sampling 100 gradient values at each period and taking the empirical variance of their norm.

\begin{figure*}[t!]
    \centering
    \begin{subfigure}{0.95\textwidth}
        \vspace{0.1in}
        \centering
        \includegraphics[width=\linewidth]{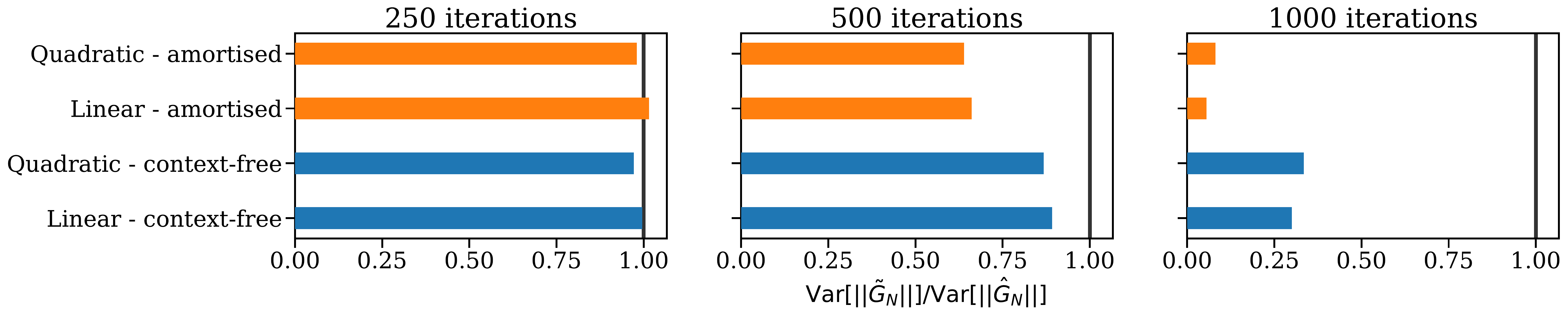}
        \caption{Logistic regression on \textit{titanic}.}
    \end{subfigure}
    \begin{subfigure}{0.95\textwidth}
        \vspace{0.1in}
        \centering
        \includegraphics[width=\linewidth]{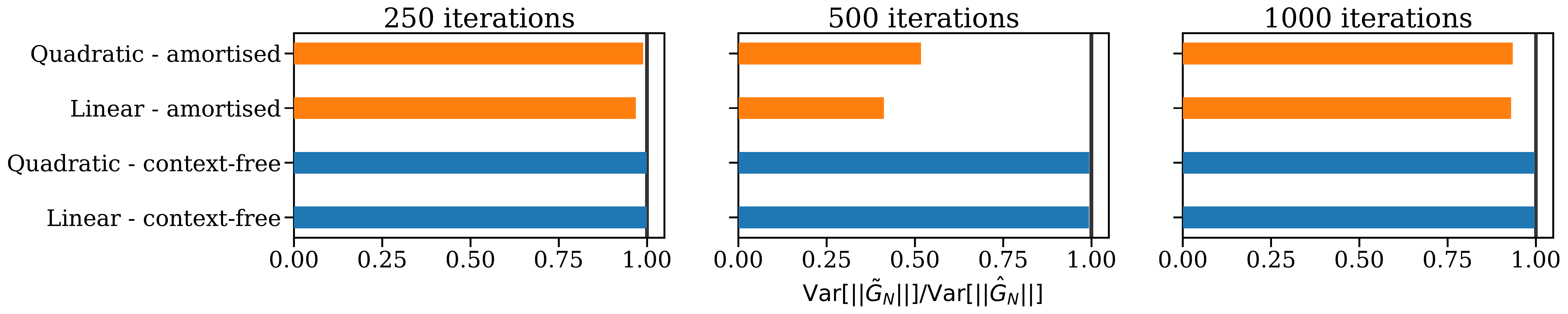}
        \caption{DGP on \textit{airfoil}.}
    \end{subfigure}
    \caption{Gradient variance ratio at three different points in joint optimisation of the model and control variate parameters (lower is better); the vertical line corresponds to a ratio of 1 (i.e. no reduction). Both the model and the control variate objectives are optimised with Adam with learning rate of $10^{-2}$ for the model and $10^{-2}$ and $10^{-3}$ for the logistic regression and DGP control variate coefficients respectively. The recognition network is able to learn a good control variate coefficient and continues to improve throughout the optimisation. The recognition network outperforms the context-free control variate.}
    \label{fig:dynamic_variance_reduction}
\end{figure*}

The recognition network is able to learn good control variate coefficients in this dynamic regime, see \cref{fig:dynamic_variance_reduction}. The variance reduction improves later on in the optimisation as observed in \cref{section:static_variance_reduction}. We again observe that the amortised control variate results in greater variance reduction than the context-free one.

\subsection{APPLICATION}
To show how our approach works in practice, we use it for training the logistic regression and DGP models. We apply the alternating optimisation procedure described in \cref{sec:dynamic_variance_reduction} on each for 2000 iterations with mini-batches size of 10. We record the mean value of the Negative Evidence Lower Bound (NELBO) from 100 MC samples for the logistic regression and 10 MC samples for the DGP at every iteration computed on the entire datasets.

\begin{figure*}[t]
    \centering
    \begin{subfigure}{0.95\textwidth}
        \vspace{0.1in}
        \centering
        \includegraphics[width=\linewidth]{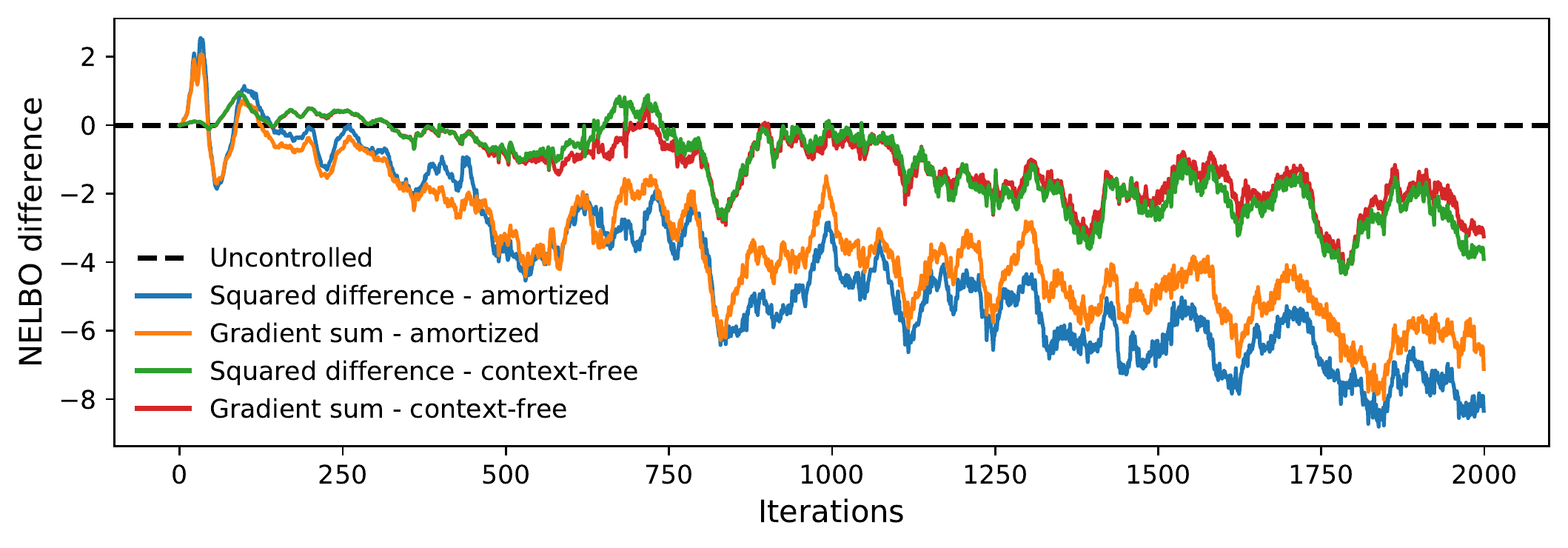}
        \caption{Logistic regression on \textit{titanic}.}
    \end{subfigure}
    \begin{subfigure}{0.95\textwidth}
        \vspace{0.1in}
        \centering
        \includegraphics[width=\linewidth]{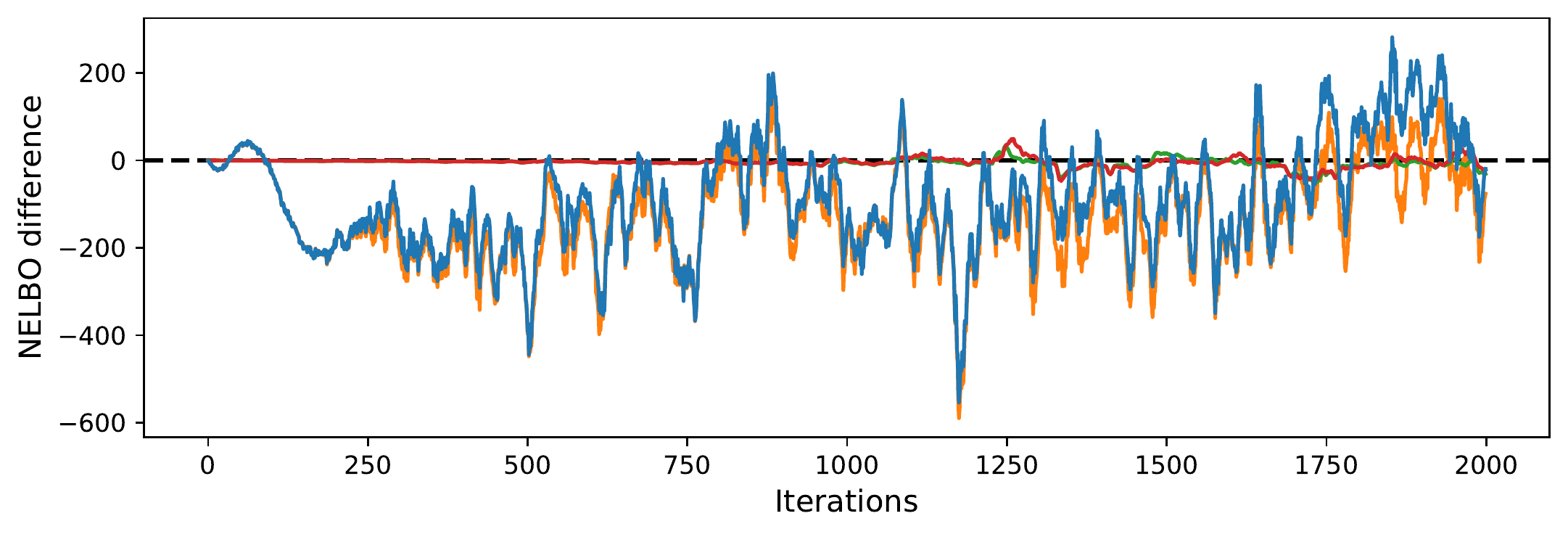}
        \caption{DGP on \textit{airfoil}.}
    \end{subfigure}
    \caption{Difference between optimisation traces for different control variate objectives, using the uncontrolled one-sample MC estimate of the gradient as a baseline (lower is better). \emph{Linear} control variates are used in this experiment. The gap between the baseline and controlled models widens through the optimisation. Amortised control variate coefficients result in wider gaps indicating better optimisation performance.}
    \label{fig:diff_trace}
\end{figure*}

The resulting traces are shown in \cref{fig:diff_trace}; in both cases we see that the optimisation with controlled gradients starts off in a worse regime than the uncontrolled gradients (curves on or above the dashed line); however, it improves as 
better control variate coefficients are learned. The gap between the one-sample MC estimator and the controlled estimators widens later for logistic regression, and fluctuates for the DGP. This is because the linear control variate sufficiently approximates the dependence of the gradient on the randomness, whereas in the case of the DGP this dependence is more complex.

For both models, amortising the control variate coefficients result in lower NELBO values on average in comparison to the uncontrolled and the context-free controlled cases. We also see that the optimisation of the control variate coefficients is robust to the choice of objective function, with similar behaviour for the gradient sum and squared difference objectives for both the amortised and context-free cases.

\cref{tab:cpu-timing} shows the average cost for the controlled optimisation steps for the two problems. Amortising the control variate coefficients with a recognition network of size [128, 128, 128] has an additional overhead of around 25\% on the context-free coefficient on the CPU. The overhead depends on many factors such as the recognition network size, control variate formulation, mini-batch size and number of gradient components. These should all be taken into account when implementing this scheme. 

\begin{table}[h]
  \vspace{1em}
  \caption{Average overall optimisation step time in \emph{milliseconds} (on the CPU) for logistic regression and DGP for different \emph{linear} control variate objective functions. The statistics are computed based on 100 repetitions of 10 runs. The implementation uses TensorFlow 2.0.}
  \vspace{1em}
  \label{tab:cpu-timing}
  \centering
  \adjustbox{max width=\linewidth}{
  \begin{tabular}{p{4cm}rr}
    \toprule
    Method & Logistic & DGP \\
    \midrule
    Squared diff. - amortised     & 1.20(0.64) & 3.77(0.22) \\
    Grad. sum - amortised     & 1.25(0.11) & 3.78(0.19) \\
    Squared diff. - context-free   & 0.87(0.91) &  3.17(0.13) \\
    Grad. sum - context-free     & 0.84(0.77) &  3.07(0.82) \\
    \bottomrule
  \end{tabular}}
\end{table}

\section{CONCLUSIONS}
We introduced a control variate formulation that exploits the structure of doubly stochastic objectives to remove Monte Carlo sampling variance from mini-batch gradient estimators. We proposed three objectives for an amortising recognition network that can learn context aware control variate coefficients. Training the network re-uses the gradients of the model objective and does not require additional passes through the model.

Empirical assessment showed that an approximation to the optimal control variate per mini-batch can be performed during optimisation and reduces the gradient variance in practice compared to a context-free global approach. In our experiments we used linear and quadratic control variates for Gaussian base randomness, but our approach is general and can be applied to other control variate formulae and randomness schemes.

\clearpage

\raggedbottom
\bibliographystyle{abbrvnat}
\bibliography{references}

\clearpage

\appendix

\onecolumn

\begin{center}
    \begin{huge}
    Supplementary Material
    \end{huge}
\end{center}

\section{VERIFICATION OF VARIANCE REDUCTION}
\begin{figure}[h!]
  \centering
  \includegraphics[width=\textwidth]{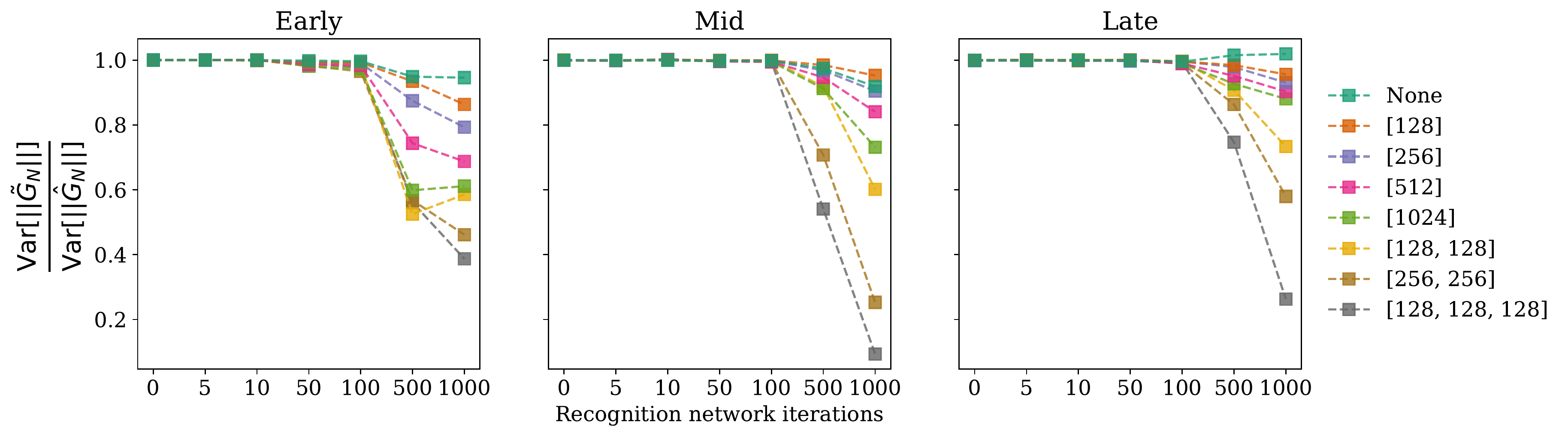}
  \caption{Logistic regression. Squared difference objective. Recognition network learning rate = $10^{-3}$.}
\end{figure}

\begin{figure}[h!]
  \centering
  \includegraphics[width=\textwidth]{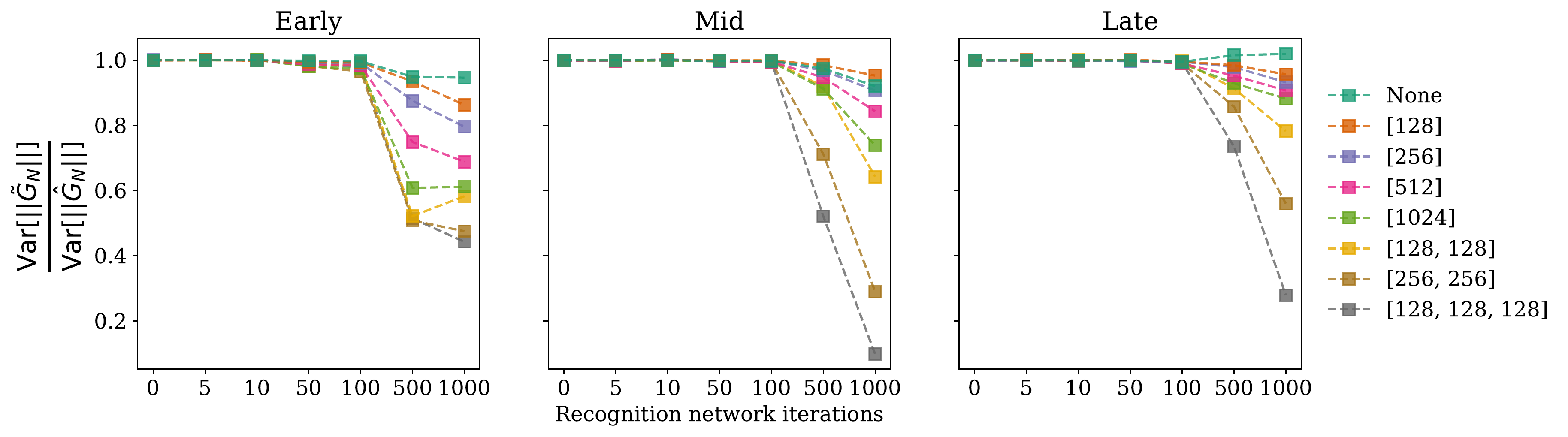}
  \caption{Logistic regression. Gradient sum objective. Recognition network learning rate = $10^{-3}$.}
\end{figure}

\begin{figure}[h!]
  \centering
  \includegraphics[width=\textwidth]{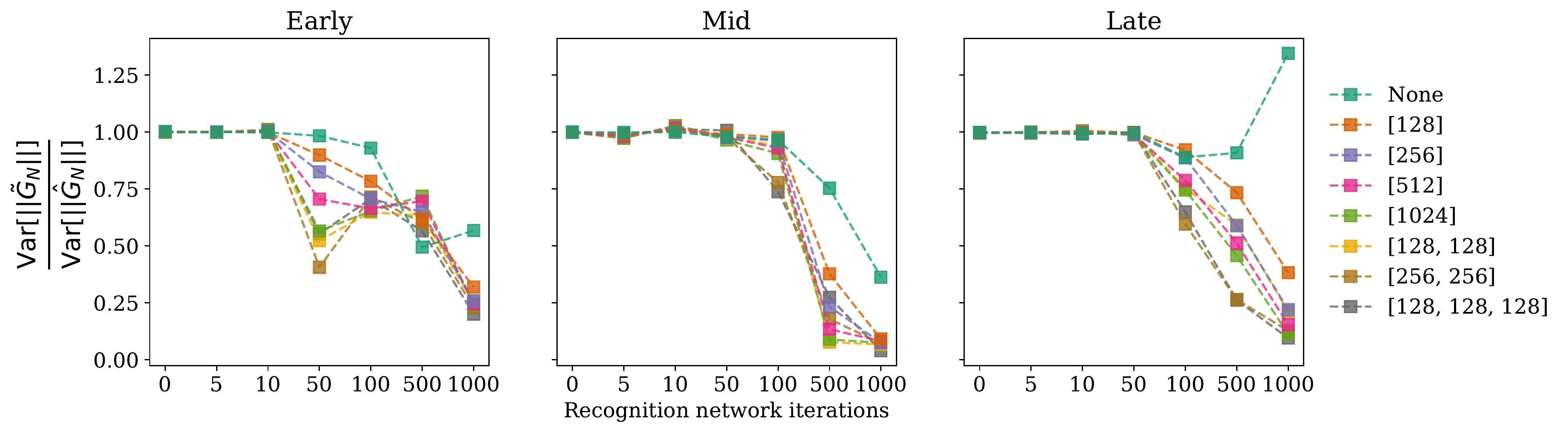}
  \caption{Logistic regression. Squared difference objective. Recognition network learning rate = $10^{-2}$.}
\end{figure}

\begin{figure}[h!]
  \centering
  \includegraphics[width=\textwidth]{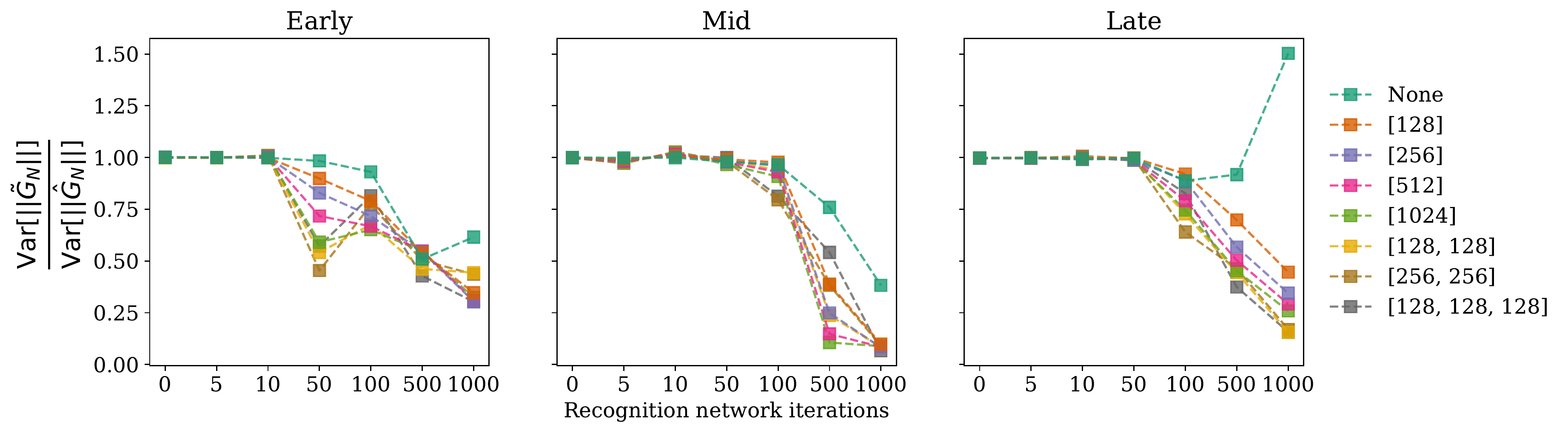}
  \caption{Logistic regression. Squared difference objective. Recognition network learning rate = $10^{-2}$.}
\end{figure}

\begin{figure}[h!]
  \centering
  \includegraphics[width=\textwidth]{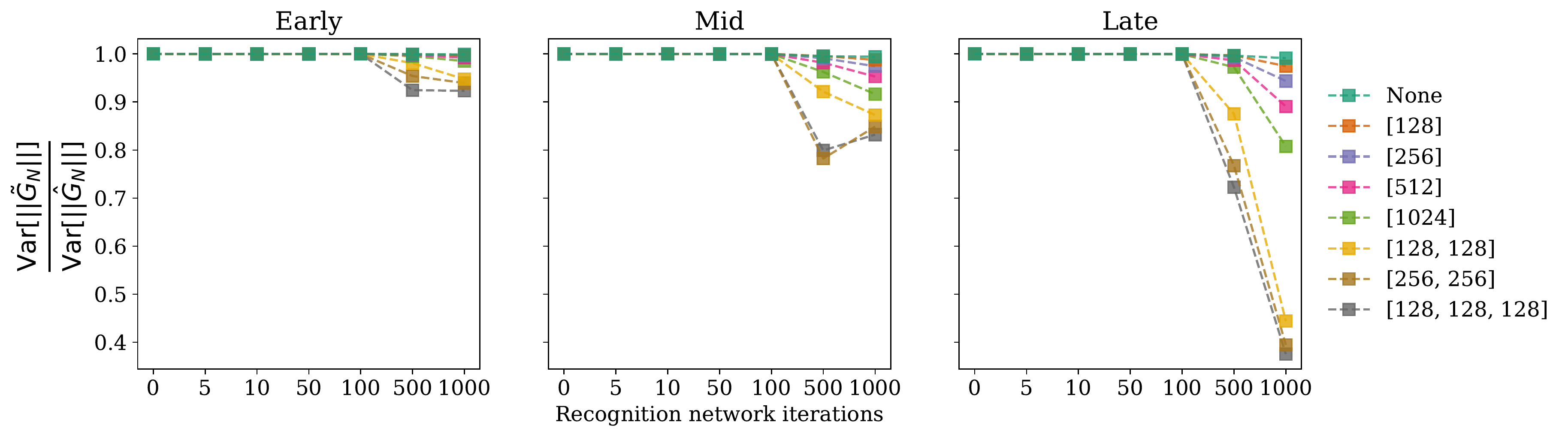}
  \caption{DGP. Squared difference objective. Recognition network learning rate = $10^{-3}$.}
\end{figure}

\begin{figure}[h!]
  \centering
  \includegraphics[width=\textwidth]{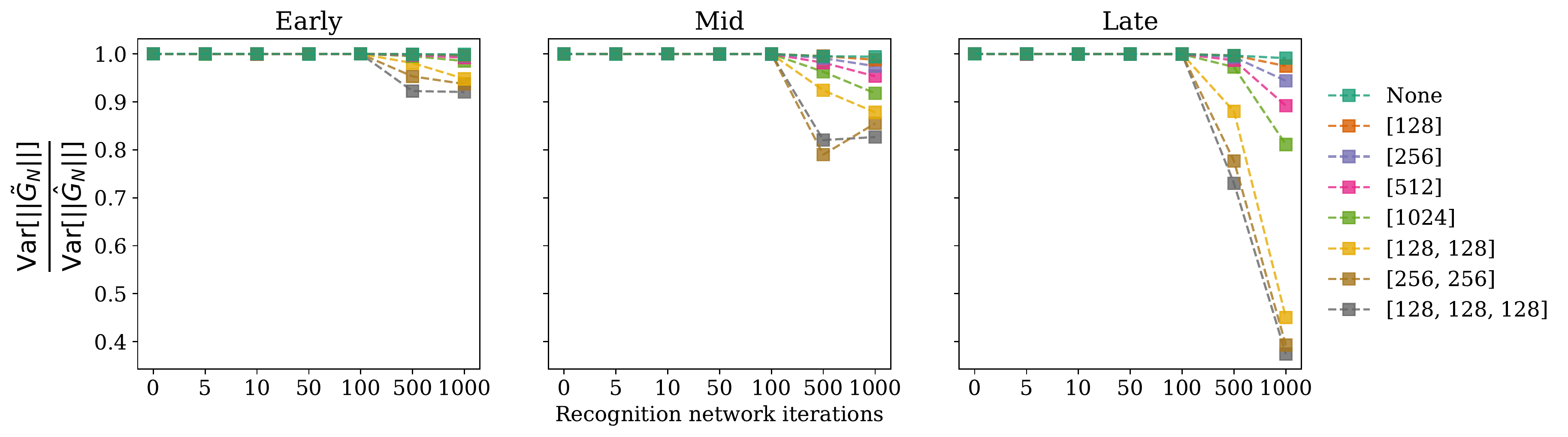}
  \caption{DGP. Gradient sum objective. Recognition network learning rate = $10^{-3}$.}
\end{figure}

\begin{figure}[h!]
  \centering
  \includegraphics[width=\textwidth]{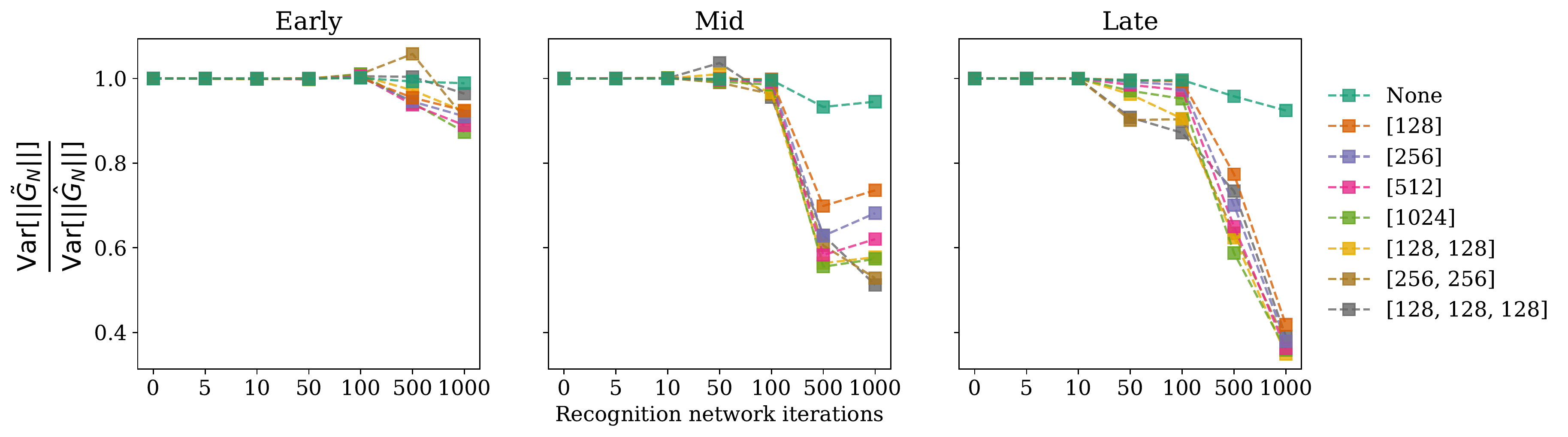}
  \caption{DGP. Squared difference objective. Recognition network learning rate = $10^{-2}$.}
\end{figure}

\begin{figure}[h!]
  \centering
  \includegraphics[width=\textwidth]{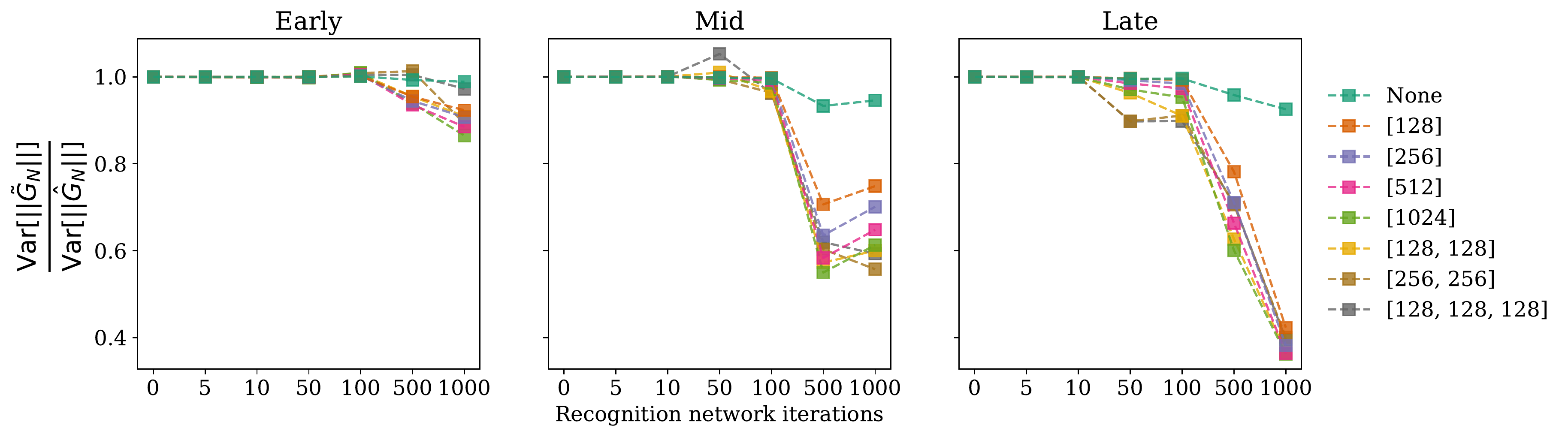}
  \caption{DGP. Gradient sum objective. Recognition network learning rate = $10^{-2}$.}
\end{figure}

\clearpage
\section{SIMULTANEOUS OPTIMISATION OF MODEL AND RECOGNITION NETWORK}

\begin{figure}[h!]
  \centering
  \includegraphics[width=\textwidth]{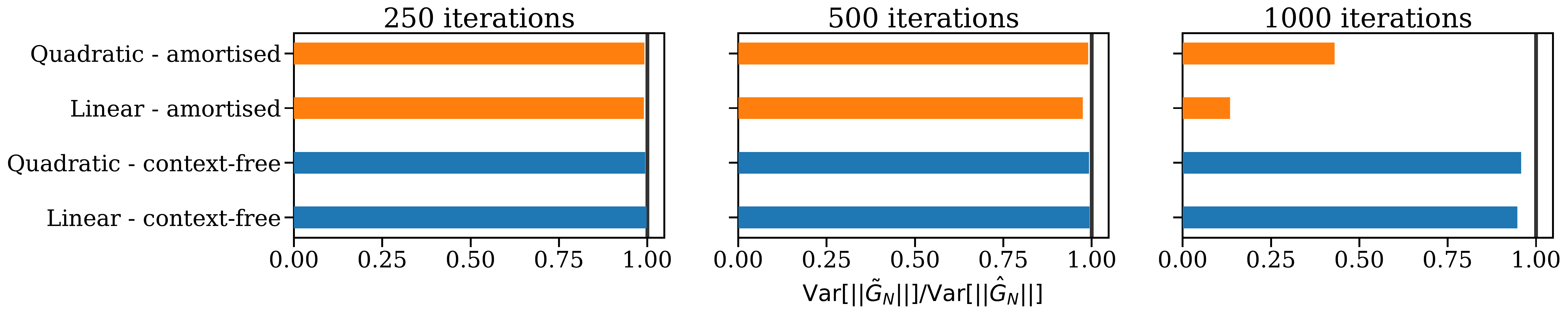}
  \caption{Logistic regression. Squared difference objective. Network of size [128, 128, 128]. Recognition network learning rate = $10^{-3}$. }
\end{figure}

\begin{figure}[h!]
  \centering
  \includegraphics[width=\textwidth]{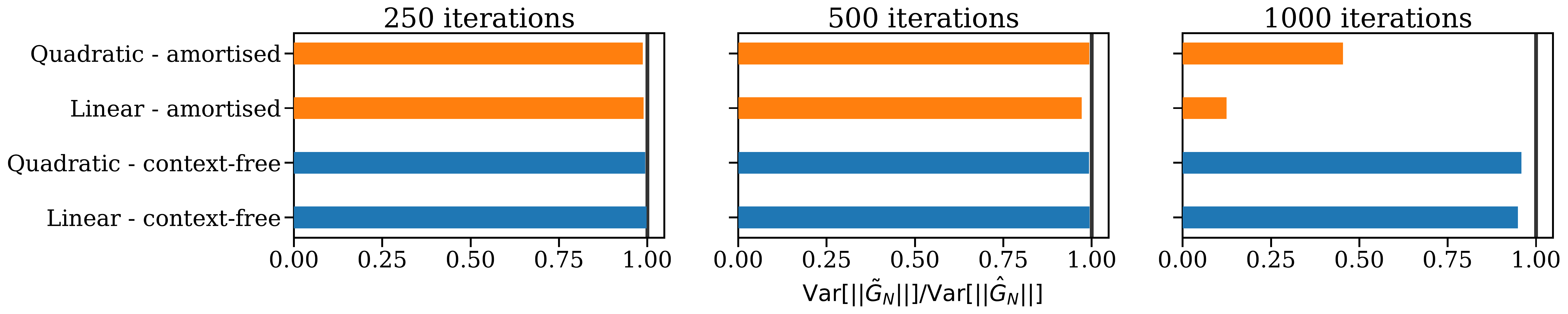}
  \caption{Logistic regression. Gradient sum objective. Network of size [128, 128, 128]. Recognition network learning rate = $10^{-3}$. }
\end{figure}

\begin{figure}[h!]
  \centering
  \includegraphics[width=\textwidth]{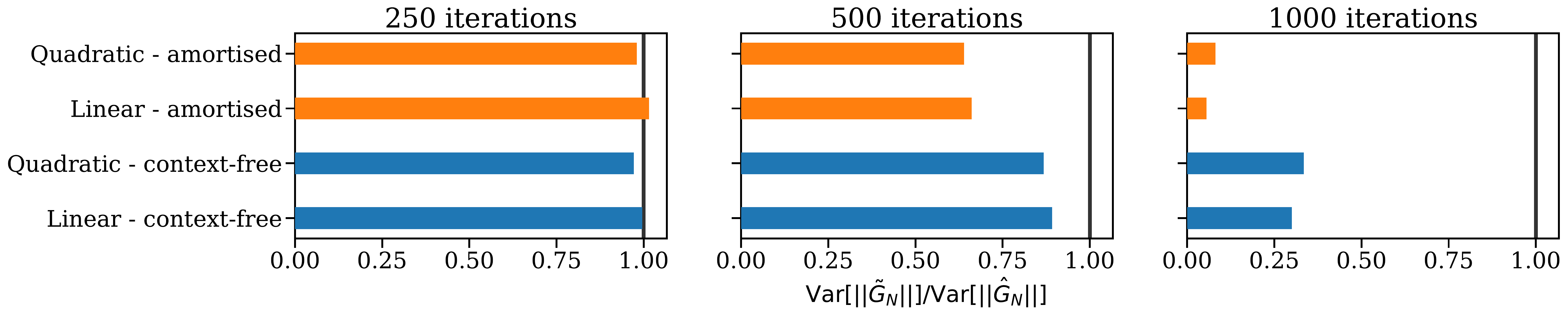}
  \caption{Logistic regression. Squared difference objective. Network of size [128, 128, 128]. Recognition network learning rate = $10^{-2}$. }
\end{figure}

\begin{figure}[h!]
  \centering
  \includegraphics[width=\textwidth]{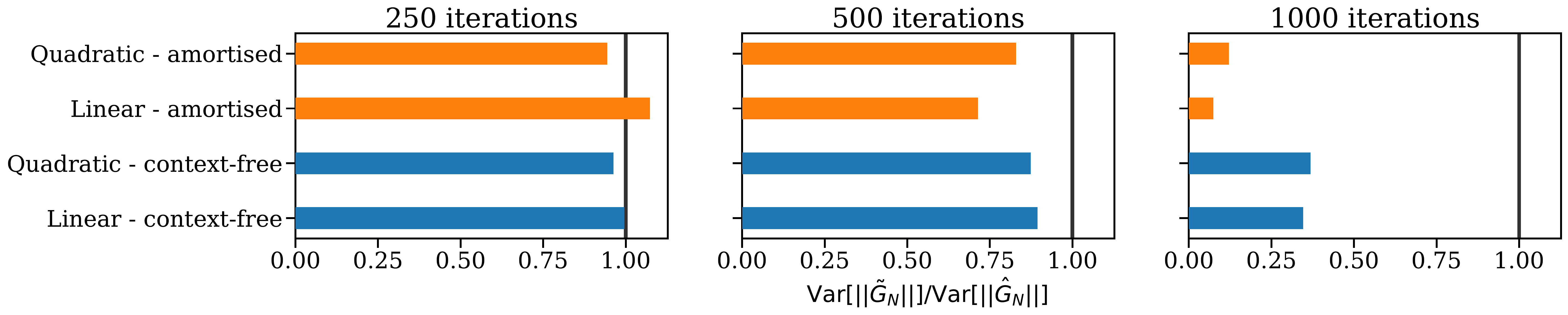}
  \caption{Logistic regression. Gradient sum objective. Network of size [128, 128, 128]. Recognition network learning rate = $10^{-2}$. }
\end{figure}


\begin{figure}[h!]
  \centering
  \includegraphics[width=\textwidth]{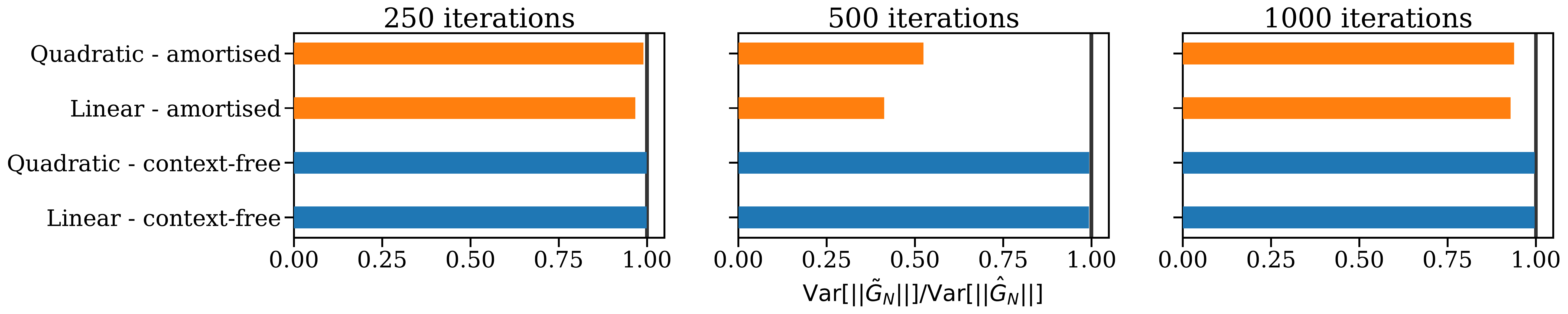}
  \caption{DGP. Squared difference objective. Network of size [128, 128, 128]. Recognition network learning rate = $10^{-3}$. }
\end{figure}

\begin{figure}[h!]
  \centering
  \includegraphics[width=\textwidth]{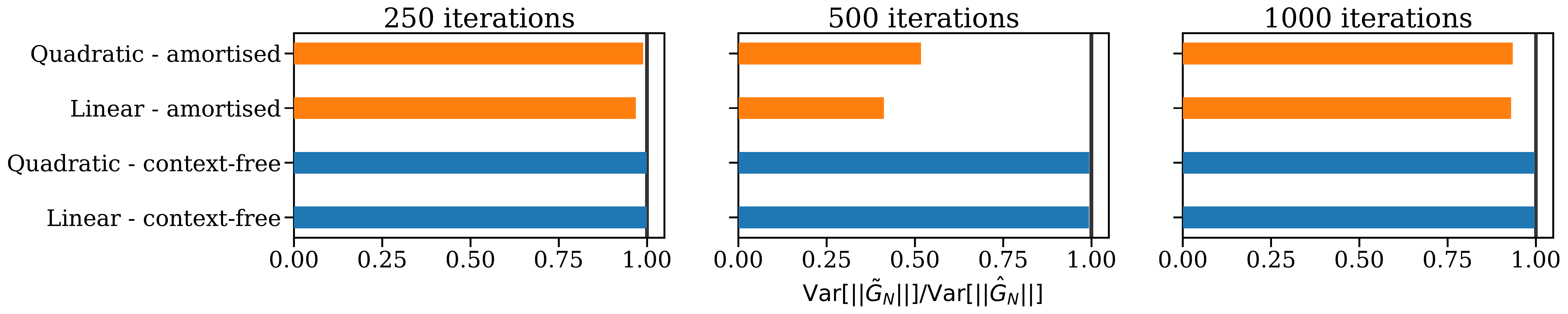}
  \caption{DGP. Gradient sum objective. Network of size [128, 128, 128]. Recognition network learning rate = $10^{-3}$. }
\end{figure}

\begin{figure}[h!]
  \centering
  \includegraphics[width=\textwidth]{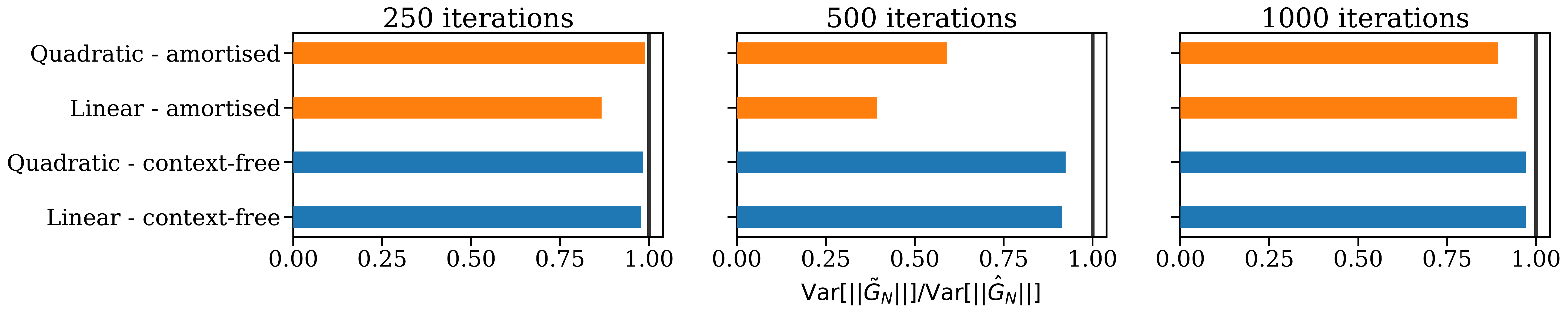}
  \caption{DGP. Squared difference objective. Network of size [128, 128, 128]. Recognition network learning rate = $10^{-2}$. }
\end{figure}

\begin{figure}[h!]
  \centering
  \includegraphics[width=\textwidth]{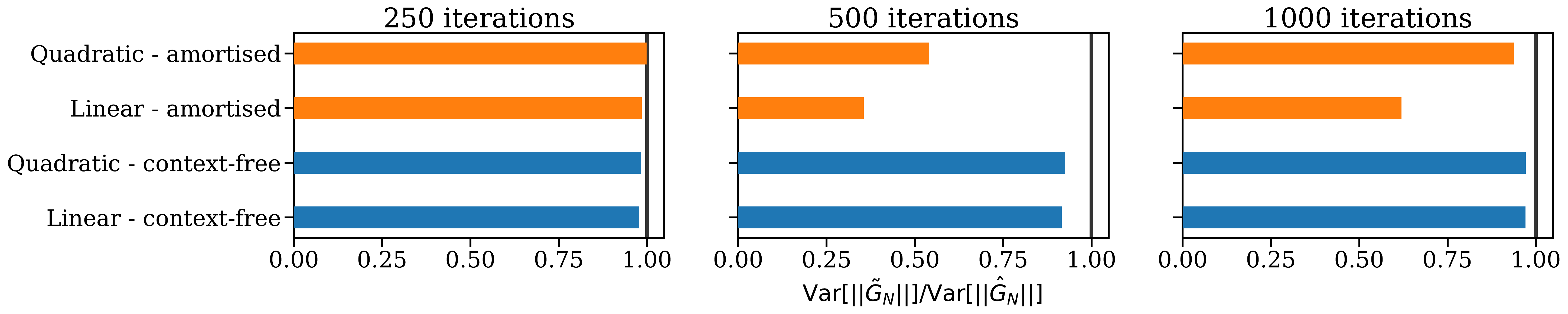}
  \caption{DGP. Gradient sum objective. Network of size [128, 128, 128]. Recognition network learning rate = $10^{-2}$. }
\end{figure}

\end{document}